\documentclass[10pt,twocolumn,letterpaper]{article}

\usepackage{cvpr}
\usepackage{times}
\usepackage{epsfig}
\usepackage{graphicx}
\usepackage{amsmath}
\usepackage{amssymb}

\newcommand*\rot{\rotatebox{90}}
\usepackage{colortbl}
\definecolor{Gray}{gray}{0.9}
\newcolumntype{g}{>{\columncolor{Gray}} c}
\usepackage{caption}

\usepackage[pagebackref=true,breaklinks=true,letterpaper=true,colorlinks,bookmarks=false]{hyperref}

\cvprfinalcopy 

\def\cvprPaperID{2595} 

\ifcvprfinal\fi
\begin{document}

\title{Adaptive Semantic Segmentation with a Strategic Curriculum of Proxy Labels}

\author{Kashyap Chitta \hspace{1cm} Jianwei Feng \hspace{1cm} Martial Hebert \\
The Robotics Institute, Carnegie Mellon University\\
{\tt\small \{kchitta,jfeng1\}@andrew.cmu.edu, hebert@cs.cmu.edu}
}


\maketitle

\begin{abstract}
Training deep networks for semantic segmentation requires annotation of large amounts of data, which can be time-consuming and expensive. Unfortunately, these trained networks still generalize poorly when tested in domains not consistent with the training data. In this paper, we show that by carefully presenting a mixture of labeled source domain and proxy-labeled target domain data to a network, we can achieve state-of-the-art unsupervised domain adaptation results. With our design, the network progressively learns features specific to the target domain using annotation from only the source domain. We generate proxy labels for the target domain using the network's own predictions. Our architecture then allows selective mining of easy samples from this set of proxy labels, and hard samples from the annotated source domain. We conduct a series of experiments with the GTA5, Cityscapes and BDD100k datasets on synthetic-to-real domain adaptation and geographic domain adaptation, showing the advantages of our method over baselines and existing approaches. 
\end{abstract}

\section{Introduction}

Dataset bias \cite{torralba2011unbiased} is a well-known drawback of supervised approaches to visual recognition tasks. In general, the success of supervised learning models, both of the traditional and deep learning varieties, is restricted to data from the domain it was trained on. Even small shifts between the training and test distributions lead to a significant increase in their error rates \cite{tzeng2017adversarial}. For deep neural networks, the common approach to handle this is fairly straightforward: pre-trained deep models perform well on new domains when they are \textit{fine-tuned} with a sufficient amount of data from the new distribution. However, fine-tuning involves the bottleneck of data annotation, which for many modern computer vision problems is a far more time-consuming and expensive process than data collection \cite{andriluka2018fluid}.

\begin{figure}[t]
\begin{center}
\includegraphics[width=0.48\textwidth]{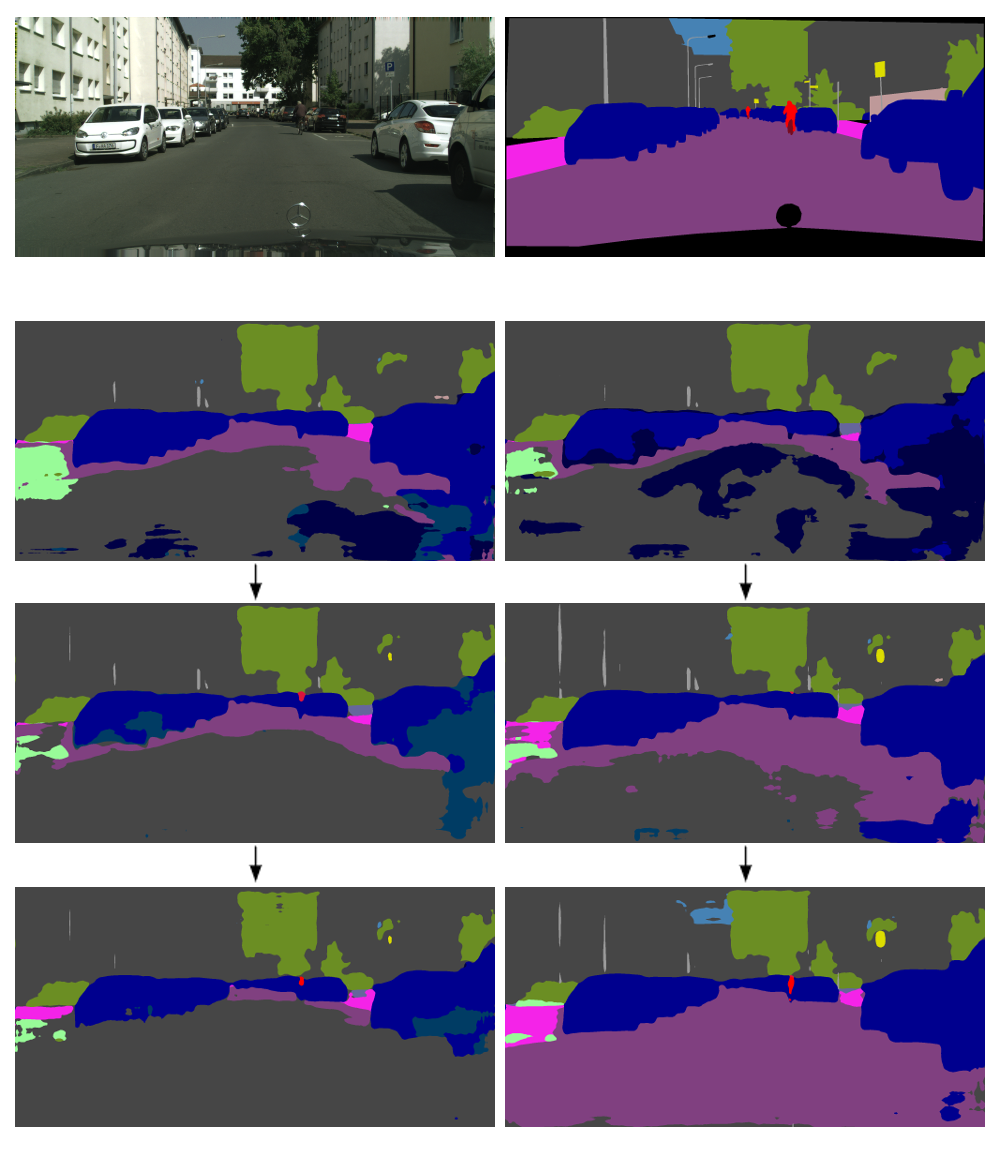}
\end{center}
   \caption{(Top row) Validation image from the Cityscapes dataset and corresponding ground truth segmentation mask. (Left column) Segmentations produced after 1, 2, and 4 epochs of self-training with GTA5 labels, where the model's mistakes are slowly amplified. (Right column) Predictions at same intervals using our strategic curriculum with target easy mining and source hard mining. The network progressively adapts to the new domain.}
\label{f:pull}
\end{figure}

Unsupervised domain adaptation aims at overcoming this dataset bias using only unlabeled data collected from the new target domain. Recent work in this area addresses domain shift by aligning the features extracted from the network across the source and target domains, minimizing some measure of distance between the source and target feature distributions, such as correlation distance \cite{sun2016coral} or maximum mean discrepancy \cite{long2015transferable}. The most promising techniques under this broad class of \textit{feature alignment} methods, referred to as adversarial methods, involve minimizing the accuracy of a domain discriminator network that classifies network representations according to the domain they come from \cite{ganin2015backprop,tzeng2015simultaneous,tzeng2017adversarial}. However, implementing these adversarial approaches requires dealing with a difficult to optimize objective function.

In this study, we explore an orthogonal idea, routinely applied while teaching people (or animals) to perform new tasks, which is to choose an effective sequence in which training examples are presented to a learner \cite{krueger2009flexible}. When applied to a machine learning system, this has been shown to have to potential to remarkably increase the speed at which learning can occur \cite{bengio2009curriculum}. In a domain adaptation setting, performing well on the target domain is hard for the network due to the lack of annotated, supervised training. We progressively learn target-domain specific features by suitably ordering the training data being used to update the network's parameters.

Our system utilizes a prominent class of semi-supervised learning algorithms that use the partially trained network to assign proxy labels to the unlabeled data and augment the training set, referred to as self-training \cite{ruder2018strong}. We show that applying a strategic curriculum while training with this augmented dataset can help overcome domain shift in semantic segmentation networks, without explicitly aligning features in any way. The strategy we use is based on easy and hard sample mining from the target and source domains, while also gradually shifting the overall distribution of data presented to the network towards the target domain over training. The method is far simpler to optimize than adversarial techniques while offering superior performance.

Our contributions are as follows: first, we implement a simple approach to perform self-training using a mixture of batches from the source and target domains with a gradient filtering strategy. Second, we design a lightweight architecture and suitable regularization scheme to evaluate the difficulty of training samples, based on prediction agreement in an ensemble of network branches. Finally, we present a strategic training curriculum to make our self-training approach better suited for unsupervised domain adaptation. We evaluate our method on several challenging domain adaptation benchmarks for urban street scene semantic segmentation, performing a detailed analysis and comparing to various baselines and feature alignment methods. We also visualize the working of our algorithm.

\section{Related Work}



\noindent
\textbf{Unsupervised Domain Adaptation.} There exist several situations where data collection in a target domain is extremely cheap, but labeling is prohibitively expensive. A prime example is semantic segmentation in new domains, as the means of collecting raw data (images or videos) is fairly cheap as opposed to the cost of annotating each pixel in this acquired data. Without annotation, fine-tuning is not possible, so the domain shift can be overcome by unsupervised domain adaptation.

Adversarial training has emerged as a successful approach for this \cite{ganin2015backprop,tzeng2017adversarial,chen2018domain}. The core idea is to train a domain discriminator to distinguish between representations from the source and target domains. While the discriminator tries to minimize its domain classification error, the main classification network tries to maximize this error term, minimizing the feature discrepancy between both domains. This is closely related to the way the discrepancy between the distribution of the training data and synthesized data is minimized in Generative Adversarial Networks (GANs) \cite{goodfellow2014gan}.

More recently, state-of-the-art unsupervised domain adaptation methods directly align the inputs between the source and target domains rather than learned representations \cite{hoffman2017cycada,murez2017image,li2018semantic,zhang2018fully,dundar2018domain}. This is achieved through the significant advances in literature on pixel-level image-to-image translation, through techniques such as cycle-consistent GANs \cite{zhu2017cycle} and image stylization \cite{li2018closed,li2018learning}. In our results, we compare against these methods. Our approach, while being simple to implement, is also orthogonal to these ideas, meaning their benefits could potentially be combined, albeit with greater difficulty in optimization and parameter tuning.

\noindent
\textbf{Self-Training.} Semi-supervised learning has a rich history and has shown considerable success for utilizing unlabeled data effectively \cite{zhu2006semi}. A prominent class of semi-supervised learning algorithms, referred to as self-training, uses the trained model to assign proxy labels to unlabeled samples, which are then used as targets while training in combination with labeled data. These targets are now no longer truly representative of the ground truth, but provide some additional noisy training signal. Most practical self-training techniques require a better way to utilize this noisy annotation by avoiding those data points for which the model assigning proxy labels is not very confident \cite{yarowsky1995unsupervised,mcclosky2006effective}.

Classic self-training has shown limited success \cite{reichart2007self,huang2009self,asch2016predicting,goot2017normalize}. Its main drawback is that the model is unable to correct its own mistakes. If the model's predictions on unlabeled data are confident but wrong, the erroneous data is nevertheless incorporated into training and the model's errors are amplified. This effect is exacerbated under domain shift, where deep networks are known to produce confident but erroneous classifications \cite{szegedy2013intriguing}. Our work aims to counter this by guiding the self-training process, selectively presenting samples of increasing levels of difficulty to the network as training progresses. We illustrate this effect by visualizing the predictions on a validation image in the Cityscapes dataset over different epochs of our training experiments in Fig. \ref{f:pull}. We add more detailed quantitative results comparing our method to self-training in our experiments.

Our approach is closely related to a variant of self-training called tri-training, which has been applied successfully for both semi-supervised learning and unsupervised domain adaptation \cite{zhou2005tritraining,saito2017asymmetric,zhang2017fully}. However, we incorporate a more generalized framework which is not restricted to networks with three branches, inspired by recent literature on ensemble-based network uncertainty estimation \cite{lakshminarayanan2017simple,beluch2018power,geifman2018boosting}.

\noindent
\textbf{Self-Paced and Curriculum Learning.} The core intuition behind self-paced learning is that rather than considering all available samples simultaneously, an algorithm should be presented training data in a meaningful order that best facilitates learning \cite{kumar2010self}. The main challenge is measuring the difficulty of samples. In most cases, there is no direct measure of this, as what is difficult to the network may change dynamically as the network is being trained. There is existing work on estimating difficulty through a teacher network \cite{thangarasa2018self,kim2018screenernet} or the assigning difficulty weights to different samples, and then sorting by these \cite{jiang2014easy,graves2017automated}. Our work, though not directly a form of curriculum learning, can be seen as a simplified version of difficulty weighting methods. We assign all samples to certain difficulty levels, which are then used to facilitate the transfer from easier to harder samples over training.

\section{Method}

\subsection{Self-Training for Domain Adaptation}

Our approach is based on self-training: we generate labels for unlabeled data in the target domain by using the network's own predictions.

\noindent
\textbf{Source-Target Ratio.} One intuition behind our method is that the representations learned by a network trained in a certain source domain may become more difficult to adapt to the target domain if the source network is fully optimized till convergence. Instead, we attempt to begin the adaptation of features before they become domain-specific, by training the network with a mixture of input data from both domains. Over the course of training, we present data in batches, based on a source-to-target batch ratio $\gamma$. Initially, the network observes more batches from the source domain. As training progresses, the number of target domain samples is increased, and finally maximized towards the end of training.

\noindent
\textbf{Gradient Filtering.} A typical curriculum based training setup would involve presenting samples to the network in a specific order, which is determined in advance through a measure of sample easiness or difficulty.

We propose a different approach, to modify our source-target ratio based sampling of data to incorporate information about the sample difficulty. Our strategy filters out the gradients of some hard samples in the early stages of training, and other easier samples during the later stages. This, in effect, allows us to determine the order in which samples are used for updating the network weights, even though batches are presented to the network by random sampling.

\begin{figure*}[t]
\begin{center}
\includegraphics[width=\textwidth]{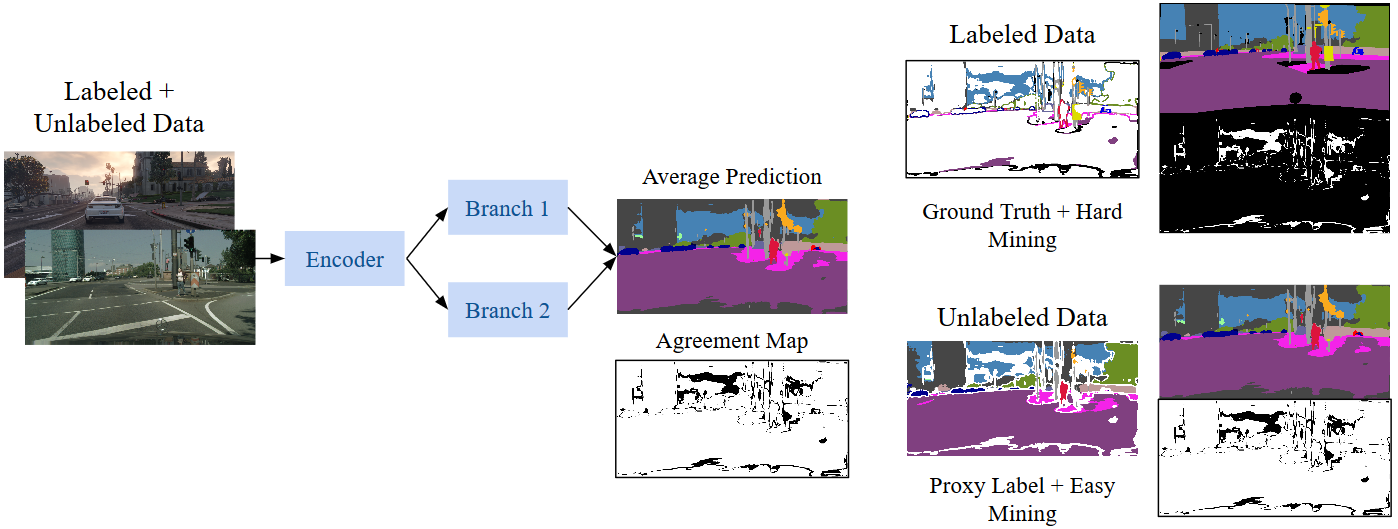}
\end{center}
   \caption{Summary of our training procedure. Batches of data, both labeled and unlabeled, are input to the network. Using two segmentation branches, we obtain an average prediction map, and an agreement map splitting easy from hard pixels. We randomly drop easy pixels while making gradient updates with the labeled data. For the unlabeled data, the network's predictions on the easy pixels are used as targets for calculating the loss. }
\label{f:method}
\end{figure*}

\subsection{Adaptive Segmentation Architecture}

Segregating the predictions based on the uncertainty in the samples is the most essential component of our training strategy. To facilitate this, our learner is constructed in a way that produces multiple segmentation hypotheses for each input image, using a network ensemble. 

\noindent
\textbf{Ensemble Agreement.} A successful technique for approximating network uncertainty in recent literature is through \textbf{variation ratios} in model ensembles \cite{beluch2018power}. This refers to the number of non-modal predictions (that do not agree with the mode, or majority vote) made by an ensemble of networks trained with different initial random seeds,

\begin{equation}
    M = \underset{e \in E}{\text{Mode}}(\underset{k \in K}{\arg \max \ } \textbf{p}_k^{(e)})
\end{equation}

\begin{equation}
f_m = \sum_{e \in E} (\underset{k \in K}{\arg \max \ } \textbf{p}_k^{(e)} = M)
\end{equation}

\begin{equation}
    VR = 1 - \frac{f_m}{E}
\end{equation}

Where $VR$ represents the variation ratios function, $M$ is the mode of the predictions and $f_m$ is the frequency of the mode. In a semantic segmentation setup, $\textbf{p}_k^{(e)}$ refers to the normalized probability output by the network indexed by $e$ in the ensemble at a given pixel location, for class $k$. $E$ is the set of all networks in the ensemble, and $K$ is the set of all classes. High variation ratios correspond to large disagreements among ensemble members, which are likely to be harder samples than those where all the ensemble members are in agreement.

\noindent
\textbf{Branch Decoder Ensembles.} If an ensemble of networks is optimized \textit{in parallel}, variation ratios can be obtained at any particular iteration during the optimization. However, this requires a significant increase in computational resources, especially on semantic segmentation networks which have huge memory footprints. Lower batch sizes or input resolutions are required to reduce this footprint, which may cause a large negative impact on the overall segmentation performance. We propose two simplifications to significantly bring down the memory requirements while retaining the ability to measure variation ratios.

First, we only ensemble \textit{a part of the network architecture}, specifically, the decoder that produces the final output from an encoded feature map. This leads to a branched architecture with a shared encoder and multiple decoders. Most segmentation architectures concentrate heavy processing in the encoder and have relatively lightweight decoder layers. Only ensembling the decoders leads to a much smaller overhead in terms of parameters, computation and memory consumption.

Further, we employ \textit{only two decoder branches} in our experiments. The number of decoders quantizes the number of levels of difficulty by which we sort our data. With two branches, variation ratios become a binary measure of agreement or disagreement, sorting samples into easy and hard pixels. This simplification is motivated by the fact that similar binary measurements of difficulty have been fairly successful in tri-training approaches to semi-supervised learning \cite{zhou2005tritraining}. 

\noindent
\textbf{Decoder Similarity Penalty.} If the two decoders learn identical weights, agreement between them with respect to a prediction would no longer be characteristic of the sample difficulty. In ensemble based uncertainty methods, different random initialization for the ensemble members has been shown to cause sufficient diversity to produce reliable uncertainty estimates \cite{lakshminarayanan2017simple}. However, in our case, we are ensembling far fewer parameters, only 2 branches of the same network, and would like to enforce diversity between the branches making segmentation predictions. This helps avoid a collapse back into regular self-training.

To facilitate this, we introduce a regularization penalty to our training objective based on the cosine similarity between the filters in each decoder. We obtain the filter weight vectors $\textbf{w}_1$ and $\textbf{w}_2$ by flattening the weights of both decoders and concatenating them.

\noindent
\textbf{Summary.} Fig. \ref{f:method} gives a full summary of the network architecture and training procedure. The network loss is calculated by independently adding up the loss at two segmentation branches during training, along with the similarity penalty, as follows:

\begin{equation}
\sum_{i \in F} l(y^i,\mathcal{B}_1(\mathcal{E}(\textbf{x}^i))) + l(y^i,\mathcal{B}_2(\mathcal{E}(\textbf{x}^i))) + \beta \textbf{w}_1^T \textbf{w}_2,
\end{equation}

where $\{(\textbf{x}^i,y^i)\}_{i=1}^{N}$ is the training data, $F$ is the subset of this training data which we retain after gradient filtering, $l$ is the cross-entropy loss for classification, and $\mathcal{E}, \mathcal{B}_1, \mathcal{B}_2$ denote the forward pass functions through the Encoder, Branch 1 and Branch 2 respectively. The regularization penalty is weighed by the tunable hyper-parameter $\beta$.

While obtaining our validation metrics, we generate an average prediction from both decoders before applying the softmax normalization. The agreement map between the decoders serves as a measure of the variation ratios.

\subsection{Strategic Curriculum}

For adaptive semantic segmentation, we define a strategic training curriculum by adding three components to our basic self-training setup: a weighted loss, target easy mining and source hard mining.

\noindent
\textbf{Weighted Loss.} Semantic segmentation typically involves a heavy class imbalance in the training data, leading to poor performances on minority classes. This effect is further exaggerated in a self-training setup, where any bias in predictions towards majority classes in the dataset can have a large impact on performance, since these predictions are used as proxy labels for further training. 

We use a loss weighting vector $\lambda$ to assign different weights to each class in our predictions to help counteract this effect. Typically, median inverse frequency based approaches are used for weighting, but we find the calculated class weighting has extreme values, destabilizing the training. Instead, we choose specific classes for which the loss weight is doubled, based on validation set performance.

\noindent
\textbf{Target Easy Mining.} For gradient filtering, the first assumption we make is that gradients from hard target domain samples may be unsuitable for updating the network. When both the decoders are unable to find consensus regarding the class prediction at a pixel, one of the two decoders is definitely making an incorrect prediction. By masking out the self-training gradients for these pixels, we avoid a large number of erroneous weight updates.

\noindent
\textbf{Source Hard Mining.} Studies in tasks with class imbalance have shown how gradients can be overwhelmed by the component coming from well-classified samples \cite{lin2017focal}. We would like to minimize the data being used to train the network from the source domain during later stages of training, and propose to do this by removing the gradients coming from these well-classified samples. 

We achieve this by randomly masking gradients of easy pixels from the source domain. This leads to us {choosing more hard pixels}, where one (or both) of the decoders made a classification error. The network is allowed to consistently improve its performance in the source domain but minimize the number of source domain training images used for weight updates. In this way, we make the features less domain-specific.

\section{Experiments}

\subsection{Datasets}

We experiment with three datasets for semantic segmentation: the synthetic GTA5 dataset \cite{richter2016playing}, and real benchmarks Cityscapes \cite{cordts2016cityscapes} and BDD100k \cite{yu2018BDD}. This lets us analyze domain shifts of varying kinds (simulated to real, geographic) and complexity. The task on which we evaluate, common across all three datasets, is street scene semantic segmentation. There are 19 compatible labeled classes over the three datasets.

\noindent
\textbf{GTA5.} This dataset contains 24,966 images of resolution $1914 \times 1052$, extracted from photo-realistic renderings of the open-world computer game Grand Theft Auto V. This is a fairly large number for a semantic segmentation dataset, highlighting ease of annotation of synthetic data as opposed to real data.

\noindent
\textbf{Cityscapes.} This dataset consists of $2048 \times 1024$ fine pixel-annotated images collected from streets over various European cities in and around Germany. There are 2,993 images in the training set and 503 images for validation. Nearly all the images are collected in good driving conditions, during daytime with clear weather. Additionally, there are 20,000 images provided with annotation in the form of coarse internal polygons, that do not cover the fine boundaries between pixels from different classes. However, we do not use these additional images in our domain adaptation experiments.

\noindent
\textbf{BDD100k.} The BDD100k benchmark has 100k images of resolution $1280 \times 720$, of which a smaller subset of 10k images have full per-pixel semantic class labels. There are 7,000 training images and 1,000 validation images. The dataset is collected in the United States, under a wide variety of driving circumstances, including different times of day, weather and lighting conditions. 

\begin{table*}[t]
  \begin{center}
  \small
  \caption{Domain adaptation results on the Cityscapes dataset, showing IoU for each class, mean IoU and pixel accuracy. Best results in \textbf{bold}. Upon incorporating all 3 components of our strategic curriculum, we outperform all competing approaches.}
  \label{t:cityscapes}
  \setlength{\tabcolsep}{1.35pt}
  \begin{tabular}{l|cccccccccccccccccccgg}
    \hline
    \multicolumn{22}{c}{\textbf{GTA5 $\rightarrow$ Cityscapes}} \\
    \hline
     &\rot{road} & \rot{sidewalk} & \rot{building} & \rot{wall} & \rot{fence} & \rot{pole} & \rot{traffic light} & \rot{traffic sign} & \rot{vegetation} & \rot{terrain} & \rot{sky} & \rot{person} & \rot{rider} & \rot{car} & \rot{truck} & \rot{bus} & \rot{train} & \rot{motorcycle} & \rot{bicycle} & \rot{\textbf{mIoU}}  & \rot{\textbf{Pixel acc.}} \\ \hline
    Source only  &21.3&13.9&48.3&3.0&3.7&22.5&24.7&
16.6&78.7&3.3&56.5&54.8&8.7&79.0&7.4&8.5&0.0&9.6&1.0&24.3&56.0\\
    \hline
    FCNs in the wild \cite{hoffman2016fcns} & 70.4 & 32.4 & 62.1 & 14.9 &  5.4 & 10.9 & 14.2 &  2.7 & 79.2 & 21.3 & 64.6 & 44.1 &  4.2 & 70.4 &  8.0 &  7.3 &  0.0 &  3.5 &  0.0 & 27.1 & ---  \\
    Curriculum \cite{zhang2017curriculum}& 74.8  & 22.0 & 71.7 & 6.0 & 11.9 & 8.4 & 16.3 & 11.1 & 75.7 & 13.3 & 66.5 & 38.0 & 9.3 & 55.2 & 18.8 & 18.9& 0.0 & 16.8 & 14.6 & 28.9 & ---  \\
    FCTN \cite{zhang2018fully} &72.2&28.4&74.9&18.3&10.8&24.0&25.3&17.9&80.1&36.7&61.1&44.7&0.0&74.5&8.9&1.5&0.0&0.0&0.0&30.5&---  \\
    ROAD \cite{chen2017road} & 76.3 & 36.1& 69.6& \bf 28.6& 22.4 & 28.6&29.3&14.8 & 82.3 & 35.3 & 72.9& 54.4&\bf 17.8& 78.9& 27.7& 30.3 & 4.0  & 24.9 & 12.6 & 39.4 & ---  \\
	CyCADA \cite{hoffman2017cycada} & 79.1&	33.1&77.9&	23.4&17.3&	 32.1&\bf 33.3&	31.8& 	81.5& 26.7&	 69.0	&\bf 62.8&	14.7& 74.5	& 20.9	& 25.6	& \bf 6.9&\bf 18.8&	\bf 20.4 & 39.5	& 82.3\\
	Domain stylization \cite{dundar2018domain} & \bf 89.0 & \bf 43.5 & 81.5 & 22.1 &  8.5 &  27.5 & 30.7 & 18.9 & 84.8 & 28.3 &\bf 84.1 & 55.7 &  5.4 &  83.2 & 20.3 & 28.3 & 0.1 & 8.7 & 6.2 & 38.3 &\bf 87.2\\
	\hline
	Self-training &62.5&24.9&72.0&8.1&1.3&20.2&0.4&0.0& 84.5& 6.1&79.9&53.2&0.0& 80.5&17.2& 11.5&0.0&0.0&0.0&27.5&74.9\\
	+ Weighted Loss &77.9&23.1&70.1&23.4&6.1&29.9&3.3&0.0& 85.0& 29.0&67.3&48.1&0.0& 81.9&25.1& 0.9&0.0&0.0&0.0&30.1&81.2\\
	+ Target easy mining&84.5&34.3&\bf 82.1&23.7&18.5&33.2&24.9&28.0&\bf 86.1&\bf 40.0&77.7&56.3&2.4&\bf 85.6&23.8&\bf 44.5&0.0&0.0&0.0&39.2&85.4\\
	+ Source hard mining&86.8&42.5&82.0&21.2&\bf 28.8&\bf 35.3&30.1&\bf 36.7&84.1&39.2&76.4&45.4&14.2&85.4&\bf 28.6&37.2&0.0&0.0&0.0&\bf 40.7&86.1\\
	\hline
	VEIS instance labels \cite{saleh2018effective} &79.8 &29.3 &77.8 &24.2 &21.6 &6.9 &23.5 &44.2 &80.5 &38.0 &76.2 &52.7 &22.2 &83.0 &32.3 &41.3 &27.0 &19.3 &27.7 &42.5& ---  \\
	Target labels &96.0&71.9&88.4&32.4&47.6&43.6&42.3&57.9&89.3&57.1&91.1&70.4&33.6&88.2&34.2&53.2&44.0&20.1&64.9&59.3&93.2\\
    \hline
  \end{tabular}
  \end{center}
\end{table*}

\subsection{Experimental Setup}

We use a ResNet-18 \cite{he2016deep} backbone for our adaptive segmentation network. The final and penultimate residual blocks are dilated by factors of 4 and 2 respectively to maintain high resolution feature maps \cite{yu2016multiscale}. We incorporate Pyramid Pooling Modules (PPMs) \cite{zhao2017pspnet} as decoders, which are segmentation architectures specifically proposed to extract global contextual cues.

The PPM adds 4 stacks of representations to the encoded feature map by using adaptive average pooling. This involves pooling the deepest available feature map with kernels of varying size (based on the input resolution) while maintaining fixed output resolutions of 6$\times$6, 3$\times$3, 2$\times$2 and 1$\times$1. The number of channels in each of these new, smaller feature maps is reduced to $\frac{1}{4}$ the original value through $1\times1$ convolutional layers. They are then upsampled back to the original resolution and concatenated to the feature map from before the pooling operations.

Once the global context features have been added to the original feature map, a $1\times1$ convolutional layer makes final class predictions at each feature location, which is upsampled back to the input resolution through bilinear interpolation.

To speed up convergence, we pre-train the segmentation model using the ADE20k dataset \cite{zhou2017scene}, a scene parsing benchmark consisting of 150 semantic classes.

We implement our system on PyTorch \cite{paszke2017automatic}. To show robustness, we adopt a common set of hyper-parameter settings across all tasks, optimized for the initial GTA $\rightarrow$ Cityscapes task, though further gains could be achieved through more detailed tuning. We optimize with Stochastic Gradient Descent (SGD), with an initial learning rate $\alpha_0$ of $0.001$ for the encoder and $0.01$ for the two decoders, which is decreased based on a polynomial decay function:

\begin{equation}
\alpha = \alpha_0 (1-\frac{e}{e_{max}})^{\delta}.
\end{equation}

We use a decay factor $\delta$ of $0.9$ every epoch. We trained for a total of $e_{max}=20$ epochs on three GTX 1080 GPUs (each with $8$GB memory), with an overall batch size of $12$ ($4$ per GPU). Batch normalization statistics are not synchronized, but maintained independently on each GPU. The momentum parameter for SGD is set to $0.9$, and weight decay to $10^{-4}$. The decoder similarity penalty $\beta$ is set to $0.01$.

For data augmentation, we scale our inputs (maintaining the aspect ratio) to a random value between 0.5 and 1.5 times the image size, and randomly flip images horizontally. Random crops from these augmented inputs of size $600 \times 600$ pixels are used to build the training batches. The source-to-target batch ratio $\gamma$ is linearly reduced from its initial value of $0.9$ to $0.1$ over the first 10 epochs of training, after which it maintains its value of $0.1$ for the remainder of the training. This means that during the latter half of training, we are almost exclusively using data from the target domain.

\subsection{Domain Adaptation on Cityscapes}

We evaluate our method against several baselines in addition to existing approaches for domain adaptation on the GTA $\rightarrow$ Cityscapes task. Since Cityscapes is our target domain, the ground truth segmentation maps for these images are not utilized by our algorithm.

Networks trained with supervised learning on the source and target datasets provide a lower and upper bound for the expected domain adaptation results. We apply naive self-training, followed by the introduction of three different components of our algorithm in steps: weighted loss, target easy mining and finally source hard mining.

For weighting the loss, we set the vector $\lambda$ so as to double the weight of the following classes: \textit{wall, fence, pole, traffic light, traffic sign, rider}. We found that choosing only classes with sufficiently large numbers of instances in the dataset but relatively poor self-training performance gave the best results.

To implement target easy mining, we simply drop all pixels where the decoders disagree from self-training. For source hard mining, we randomly drop a portion of the pixels where the decoders agree in the source domain. This ratio is varied linearly from an initial value of $0$ to a final value of $1$ after 10 epochs. Similar to $\gamma$, this value is then retained for the final 10 epochs of training.

\begin{figure}[t]
\begin{center}
\includegraphics[width=0.45\textwidth]{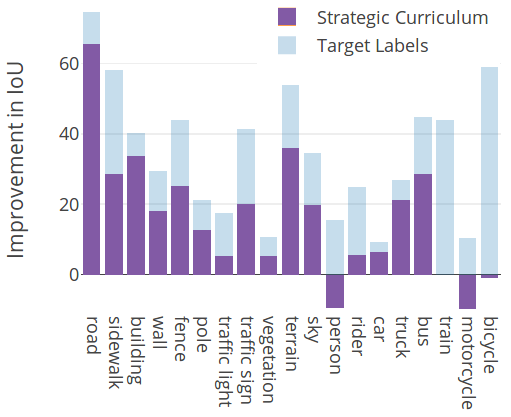}
\end{center}
   \caption{Class-wise IoU improvements through our adaptive curriculum. On most classes, we recover a large portion of the performance lost by domain shifts, however, we are unable to do so on a small subset of the classes (\textit{person, train, motorcycle, bicycle}).}
\label{f:classIoU}
\end{figure}

\begin{figure*}[t]
\begin{center}
\includegraphics[width=\textwidth]{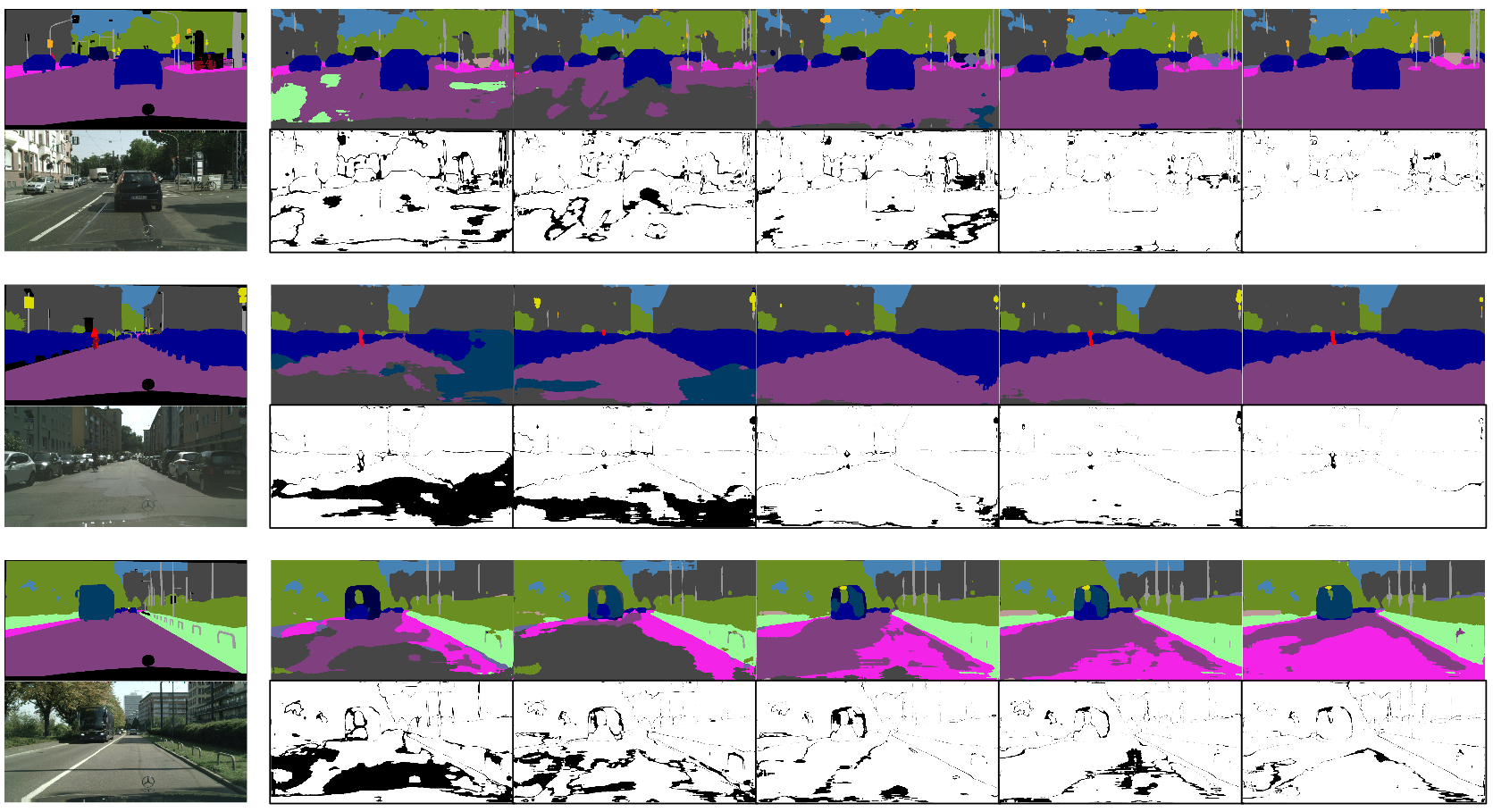}
\end{center}
   \caption{Predicted labels and agreement maps for Cityscapes validation images over different epochs of domain adaptation from GTA5. The first two sequences show the gradual improvement in network confidence through the reduction in the number of hard pixels in the segmentation map. The final sequence shows a failure case, where predictions in hard regions are incorrectly adapted.}
\label{f:vis}
\end{figure*}

\noindent
\textbf{Results.} Our results are summarized in Table \ref{t:cityscapes}. We observe a significant gap in performance between the source only baseline, with a mean IoU of 24.3, and the network trained with target labels, which is at 59.3. The worst classes, such as \textit{train} and \textit{bicycle}, are severely underrepresented in the GTA5 dataset. In addition, even the well-represented classes, such as \textit{road}, have notably poorer performance when training is carried out with only synthetic data.

Through naive self training, we obtain the expected effect of self-reinforcement. Performance on the majority classes improves (such as \textit{road}, \textit{building}, and \textit{sky}) at the expense of the other classes, as the model is unable to correct its errors. Overall, this component of our curriculum is responsible for the most significant recovery of the domain shift in terms of pixel accuracy (nearly a $20\%$ improvement) but comes with a relatively small improvement in mean IoU (around $3\%$). 

When we apply a weighted loss, we start addressing the class imbalance issues, further improving the mean IoU by a modest amount. However, the most important aspect of our approach in terms of performance gained comes next, which is easy mining in the target domain. Classes like \textit{traffic light} and \textit{traffic sign}, which through naive self-training become progressively worse, are instead improved in performance as the erroneous gradients are filtered out. This is further improved by incorporating source hard mining, taking the final pixel accuracy of our approach to $86.1\%$, a $40\%$ improvement from the baseline; and mean IoU to $40.7$, which is more than a $16$ point improvement. Hard mining leads to a large improvement in classes with poor performance (like \textit{fence} and \textit{truck}), but this happens at the expense of two of the classes with higher initial performance (\textit{bus} and \textit{person}). This indicates that some form of class-weighting within the hard mining process could be beneficial.

\begin{table*}[t]
  \begin{center}
  \small
  \caption{Domain adaptation results on the BDD100k dataset, showing IoU for each class, mean IoU and pixel accuracy. The strategic curriculum outperforms a self-training baseline on both tasks. The adaptation has greater success from synthetic to real data, where we observe large improvements and a better final mean IoU than geographic adaptation.}
  \label{t:bdd}
  \setlength{\tabcolsep}{1.35pt}
  \begin{tabular}{l|cccccccccccccccccccgg}
    \hline
    \multicolumn{22}{c}{\textbf{GTA5 $\rightarrow$ BDD}} \\
    \hline
     &\rot{road} & \rot{sidewalk} & \rot{building} & \rot{wall} & \rot{fence} & \rot{pole} & \rot{traffic light} & \rot{traffic sign} & \rot{vegetation} & \rot{terrain} & \rot{sky} & \rot{person} & \rot{rider} & \rot{car} & \rot{truck} & \rot{bus} & \rot{train} & \rot{motorcycle} & \rot{bicycle} & \rot{\textbf{mIoU}}  & \rot{\textbf{Pixel acc.}} \\ \hline
	Self-training &83.8&11.7&70.2&4.7&12.9&24.6&0.3&0.0& 77.9& 26.8&82.5&52.7&0.0& 76.6&19.2& 27.5&0.0&0.0&0.0&30.1&84.0\\
	Strategic curriculum &87.0&33.8&72.9&14.5& 31.6&31.5& 26.8& 28.2& 67.6& 31.0&80.5&53.5&3.9& 80.2& 27.9& 53.1&0.0&0.0&0.0& 38.1&  84.3\\
	\hline
	\multicolumn{22}{c}{\textbf{Cityscapes $\rightarrow$ BDD}} \\
	\hline
	Self-training &81.6&31.9&71.8&12.6&18.2&23.9&0.0&23.1& 79.0& 31.0&77.1&50.4&0.0& 78.3&0.0&40.0&0.0&0.0& 32.4&34.3&84.2\\
	Strategic curriculum & 87.9& 39.8& 75.0& 15.3&24.6& 29.1& 0.4&23.1& 77.5& 24.2& 87.0& 53.7& 9.3& 79.6&0.0& 36.4&0.0&0.0&29.8& 36.7& 86.5\\
	\hline
	Target labels &92.6 &56.6 &82.9 &30.7 &45.7 &35.3 &30.1 &37.2 &84.6 &45.9 &94.1 &60.1 &0.0 &86.3 &42.8 &68.0 &0.0 &0.0 &14.6& 47.8& 91.7 \\
    \hline
  \end{tabular}
  \end{center}
\end{table*}

Our approach outperforms all existing methods for unsupervised domain adaptation that utilize labeled data only from the GTA5 dataset during training (in terms of mean IoU). On 9 of the 19 classes, our technique improves on the IoU of the best existing method. We are also competitive (within a $5$ point IoU) on 5 of the other classes. We observe that the CyCADA approach, which involves adversarial feature alignment, is quite complementary to our method in terms of the classe-wise performance improvements. Although these two orthogonal ideas could be combined, we leave this for future work since it may be harder to correctly tune the hyper-parameters for the resulting architecture.

Our results are also competitive with \cite{saleh2018effective}, which uses additional data from the VEIS dataset, with instance level segmentation masks, to augment the training set. This allows them to introduce sufficient training data for the classes such as \textit{train}, \textit{motorcycle} and \textit{bicycle}, which are the only classes for which their approach considerably outperforms ours. Excluding these 3 classes from evaluation would improve our mean IoU to $48.4$.

We plot the class-wise IoU improvements over the source only baseline obtained through our strategic curriculum in Fig. \ref{f:classIoU}, using the upper bound with the target labels as a reference. It can be noted that on the \textit{car} class, our domain adaptation technique approaches nearly an equivalent performance to supervised training in the target domain.

\noindent
\textbf{Agreement maps and Qualitative Analysis.} An added benefit of our approach is the ability to visualize the network's evaluation of easiness or difficulty of segmentation, through agreement maps between the two decoders. We show agreement maps along with predicted segmentations on the Cityscapes validation set in Fig. \ref{f:vis}. Over training, the number of hard pixels reduces and segmentation quality improves. We observe that the disagreement map highlights class boundaries, which are harder to segment reliably, and misclassified instances of certain minority classes. However, in the rare occasions where the network makes confident misclassifications and both decoders agree, our approach fails to correct these mistakes, which are then amplified over training.

\subsection{Domain Adaptation on BDD}
We evaluate our approach against the naive self-training baseline on the two adaptation tasks involving BDD as the target domain, GTA5 $\rightarrow$ BDD and Cityscapes $\rightarrow$ BDD. Due to the lower resolution of images in the BDD dataset, we scale the source domain images to a height of 720 pixels while maintaining their aspect ratio before training. BDD is a relatively new dataset, and to our knowledge, these domain adaptation tasks have not been studied in existing work. Our results on these tasks are summarized in Table \ref{t:bdd}.

We observe a similar trend to that of the GTA5 $\rightarrow$ Cityscapes experiments, where the proposed strategic curriculum improves performance over naive self-training. One interesting observation is that on most of the foreground classes involving vehicles and signage, domain adaptation from the synthetic GTA5 data works best; while background classes are better segmented when adapting from Cityscapes. Overall, adaptation from GTA5 achieves a slightly higher mean IoU even though the self-training baseline is considerably worse than that of Cityscapes. This indicates that our technique is well-suited to the synthetic-to-real domain adaptation task.

\section{Conclusion}

In this paper, we proposed a method to exploit unlabeled data for semantic segmentation by training with proxy labels. It combines self-training with a gradient filtering strategy using a mixture of labeled and unlabeled training data. For improving unsupervised domain adaptation, we introduce three modifications: class-wise weighting of the loss function, easy mining in the proxy-labeled target domain samples, and hard mining in the labeled source domain samples. To mine easy and hard pixels, we proposed an architecture with two decoder branches. The agreement map between these branches can be used to visualize the model's uncertainty in its predictions. Our approach obtains state-of-the-art results on unsupervised domain adaptation from the synthetic to real data. Moreover, our approach is orthogonal to existing techniques based on feature alignment, and could potentially be combined with these for further improvements in performance. We further validate our idea with strong results on additional domain adaptation tasks with the challenging BDD100k benchmark.

{\small
\bibliographystyle{ieee}
\bibliography{egbib}
}

\end{document}